%% file: acl_latex.tex
\pdfoutput=1

\documentclass[11pt]{article}

\usepackage{acl}

\usepackage{times}
\usepackage{latexsym}
\usepackage{amssymb}
\usepackage{amsmath}
\usepackage{amsthm}
\usepackage[T1]{fontenc}

\usepackage[utf8]{inputenc}

\usepackage{microtype}

\usepackage{inconsolata}
\usepackage{amsmath,graphicx}
\usepackage{tabularx}
\usepackage{amssymb}
\usepackage{amsthm}
\usepackage{cleveref}
\usepackage{pifont}
\usepackage{cancel}
\usepackage{lipsum}
\usepackage{fvextra}
\usepackage{booktabs}
\usepackage{algorithm}
\usepackage{algorithmic}
\usepackage{multirow}
\usepackage{xcolor}
\usepackage{enumitem}
\usepackage{ragged2e}
\usepackage{subcaption}
\definecolor{esc}{RGB}{0, 153, 51}

\newcommand{\READ}{\texttt{READ} }
\newcommand{\WRITE}{\texttt{WRITE} }

\input{sections/macros}

\definecolor{amber}{rgb}{1.0, 0.75, 0.0}
\title{Conversational \textsc{SimulMT}: Efficient Simultaneous Translation \\ with Large Language Models}

\author{Minghan Wang\textsuperscript{1}, Thuy-Trang Vu\textsuperscript{1}, Yuxia Wang\textsuperscript{2},\\ \textbf{Ehsan Shareghi}\textsuperscript{1}, \textbf{Gholamreza Haffari}\textsuperscript{1} \\
  \textsuperscript{1}Department of Data Science \& AI, Monash University \quad \textsuperscript{2}MBZUAI \\
  \texttt{\{minghan.wang,trang.vu1,ehsan.shareghi,gholamreza.haffari\}}@monash.edu\\
  \texttt{yuxia.wang}@mbzuai.ac.ae
}

\begin{document}
\maketitle

\input{sections/0-abstract}

\input{sections/1-intro}

\input{sections/2-background}

\input{sections/3-method}

\input{sections/4-experiment}
\input{sections/4.1-analysis}

\input{sections/5-related-work}

\input{sections/6-conclusion}

\input{sections/7-limitations}

\bibliography{anthology,simul,other}

\input{sections/8-appendix}




\end{document}

%% file: sections/macros.tex
\newcommand\vx{\mathbf{x}} 
\newcommand\vy{\mathbf{y}}

\definecolor{dgreen}{rgb}{0,0.55,0}
\definecolor{mgreen}{rgb}{0,0.7,0}

\newcommand{\myitem}{\vspace*{-1mm}\item}

%% file: sections/0-abstract.tex
\begin{abstract}
Simultaneous machine translation (\textsc{SimulMT}) presents a challenging trade-off between translation quality and latency.
Recent studies have shown that LLMs can achieve good performance in \textsc{SimulMT} tasks. However, this often comes at the expense of high inference costs and latency. 
In this paper, we propose a conversational \textsc{SimulMT} framework to enhance the inference efficiency of LLM-based \textsc{SimulMT} through multi-turn-dialogue-based decoding where source and target chunks interleave in translation history, enabling the reuse of Key-Value cache. 
To adapt LLMs to the proposed conversational decoding, we create supervised fine-tuning training data by segmenting parallel sentences using an alignment tool and a novel augmentation technique to enhance generalization.
Our experiments with \texttt{Llama2-7b-chat} on three \textsc{SimulMT} benchmarks demonstrate that the proposed method empowers the superiority of LLM in translation quality, meanwhile achieving comparable computational latency with specialized \textsc{SimulMT} models.\footnote{Our code is available at \url{https://github.com/yuriak/LLM-SimulMT}}
\end{abstract}

%% file: sections/1-intro.tex
\section{Introduction}

Simultaneous machine translation (\textsc{SimulMT}) systems provide real-time translation of text input stream \cite{gu-etal-2017-learning}. This task plays an important role in real-world applications, such as facilitating communication in online conferences and generating live subtitles with strict latency requirements. 

\input{figures/fig-prompt_kvcache}

Although large language models (LLMs) have shown the potentials in machine translation~\citep{DBLP:journals/corr/abs-2302-09210,DBLP:journals/corr/abs-2304-04675}, their applications to \textsc{SimulMT} is non-trivial, as they are not inherently designed for simultaneous decoding. Recent works have attempted to adapt LLMs for \textsc{SimulMT} with prefix fine-tuning, incremental decoding~\citep{wang2023simultaneous} and learning to wait for more source tokens before translation~\citep{koshkin2024transllama}. 
These works show LLMs, with careful prompt-engineering, could approach the performance of specialized \textsc{SimulMT} models. However,
high computational cost, slow inference, and high latency render these approaches impractical for real-world applications~\citep{yuan2024llm}.
This is primarily due to the use of \emph{offline prompting}, where
arriving source tokens are inserted at the end of the source sequence, disrupting the continuity of the translation history (\Cref{fig:prompt_kvcache} left). This prevents reusing cached target history states and requires re-computation of source and target representations.

To mitigate this issue, we propose \emph{conversational prompt} that resemble the multi-turn dialogue nature of LLMs. Specifically, user inputs are treated as the source tokens to be read, while the LLM's responses are considered the predicted target tokens to be written. In our conversational \textsc{SimulMT}, newly arrived source form the current instruction, while previous source tokens and their translations are treated as conversation history (\Cref{fig:prompt_kvcache} right).
This conversational prompt enables the reuse of
Key-Value cache \cite{kv-cache2023}, as all content is appended incrementally without modifying the translation history. 
However, conversational \textsc{SimulMT} poses new challenges for LLMs to comprehend the segmented source content and produce a coherent translation via multi-turn conversation.

To adapt the LLM to the conversational decoding format, we opt to perform supervised fine-tuning (SFT) on the pretrained LLM. But the challenge is the lack of the conversational \textsc{SimulMT} data for SFT. Interleaving incomplete source and target segments in the dialogue history is unnatural (see \Cref{fig:prompt_kvcache}). This code-switching style is exhibited in some languages~\citep{yong2023prompting}; however, it is the \emph{continuation} rather than the \emph{translation} of the previous content, making it challenging to leverage existing code-switched datasets for training. Therefore, we propose to curate the training data by segmenting parallel sentence pairs into smaller chunks based on a transformation of the word alignments.
The segmented chunks are further augmented to handle different latency requirements.

Experiments on three \textsc{SimulMT} benchmarks demonstrate the effectiveness of our proposed conversational \textsc{SimulMT} in balancing the trade-offs between accuracy, speed and flexibility to different latency requirements. Compared to offline prompting, our method not only maintains strong performance, but also benefits from reduced latency. Notably, our method attains similar decoding speed to the LLM-based \textsc{OfflineMT}.

Our contributions are summarized as follows,
\begin{itemize}
    \item We introduce conversational prompting to reduce the inference cost of LLM-based \textsc{SimulMT} by leveraging its multi-turn dialogue capability and enabling efficient reuse of Key-Value cached computations.
    \item We present an automated training data curation pipeline that can turn any offline translation parallel corpus into the conversational prompt format and generalize with a novel augmentation strategy into any inference setting.
    \item Experiments demonstrate that the proposed conversational \textsc{SimulMT} obtains up to 2$\times$ acceleration compared to the offline-prompting baseline while maintaining comparable translation quality, emphasizing its value in practical applications.
\end{itemize}

%% file: figures/fig-prompt_kvcache.tex

\begin{figure}[t]
    \centering
    \includegraphics[width=0.8\columnwidth]{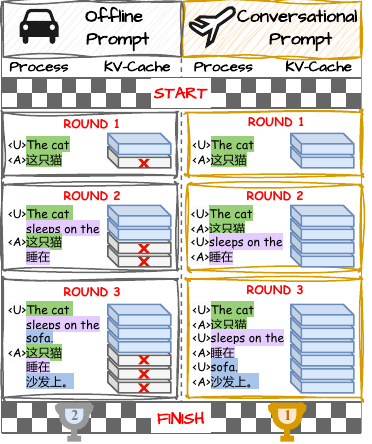}
    \caption{Comparison of offline prompt (left) and conversational prompt (right). Offline prompt inserts tokens mid-sequence, preventing KV-cache reuse (\textcolor{red}{red X}), while conversational prompt appends content sequentially, enabling efficient cache utilization (\textcolor{cyan}{blue} blocks).}
    \label{fig:prompt_kvcache}
\end{figure}

%% file: sections/2-background.tex
\section{Background}
\paragraph{Simultaneous Machine Translation (\textsc{SimulMT})} Unlike \emph{offline} machine translation (\textsc{OfflineMT}), where models generate target translation $\vy = (y_1,...,y_J)$ given a complete source sentence $\vx = (x_1,...,x_I)$, \textsc{SimulMT} incrementally translates with partial source context $\vx_{\leq t} = (x_1,...,x_t)$ where $t \leq I$. A core component of \textsc{SimulMT} is a read-write policy that decides whether to wait for new source tokens (\texttt{READ}) or generate target tokens (\texttt{WRITE}), balancing translation quality and latency.

\paragraph{Incremental Decoding}
Studies have explored adapting \textsc{OfflineMT} models for simultaneous decoding by performing offline decoding on incrementally updated histories~\citep{low_latency,nguyen21c_interspeech,polak-etal-2022-cuni,HW_TSC}. This involves a chunk-wise \texttt{READ} policy that reads $n$ tokens per round and a \texttt{WRITE} policy that commits stable partial translations using the longest common prefix (LCP)~\cite{polak-etal-2022-cuni} algorithm. LCP often causes high latency when candidates lack common initial tokens. Relaxed Agreement LCP (RALCP)~\citep{wang2023simultaneous} was proposed to vote for accepting prefixes with candidate agreement above threshold $\gamma$.

\paragraph{\textsc{SimulMT} with LLMs}
Since incremental decoding essentially repeats offline decoding, using offline-style translation prompts with LLMs is straightforward and aligns with their instruction-following capabilities~\cite{xu2023paradigmshift}. During each round, a source chunk is \READ and appended to source history $\vx$. LLMs generate translations using offline prompts as shown in~\Cref{fig:prompt_kvcache}, which are then \texttt{WRITTEN} to target history $\vy$.

%% file: sections/3-method.tex
\section{Conversational \textsc{SimulMT}}
While incremental decoding with offline prompt enables LLMs to perform simultaneous decoding, it faces high computational latency due to the insertion of newly arrived source tokens in the middle of the prompt, disrupting the reuse of cached target history states. In this section, we propose conversational prompts to improve the decoding efficiency and balance quality-latency trade-off.

\subsection{Decoding with Conversational Prompt}
\label{sec:conv_prompt}

The efficiency improvement in LLMs hinges on maintaining the Key-Value (KV-) cache reuse, i.e. the decoding process must consistently add new tokens at the end of the sequence without altering the middle elements. When LLMs are performing multi-turn dialogues, the prompt for each turn is composed of a user input and assistant response
separated by special tokens, and conversation histories are simply concatenated as the context \cite{touvron2023llama2}. Drawing parallels to multi-turn dialogues in LLMs, \textsc{SimulMT} can also be viewed similarly, where user inputs and assistant responses are equivalent to \READ and \WRITE action. At round $t$, LLM reads a source context chunk $X_t$ and writes its translation $Y_t$: ``\texttt{<U> $X_{t}$ <A> $Y_{t}$}". The already processed chunks are concatenated as contexts, serving the latest translation round of new incoming chunks. As all contents are appended incrementally, the reuse of KV-cache becomes feasible again like in multi-turn dialogue (see~\Cref{fig:prompt_kvcache}).
Our approach also adapts the hypothesis selection strategy e.g.  RALCP~\citep{wang2023simultaneous} to prune the unstable suffixes in each response. \Cref{algo:simul_decode} in \Cref{sec:appx-algo} presents the detailed decoding process.

\input{tables/tab-preliminary}

We conducted a pilot experiment to assess LLMs' zero-shot and few-shot capabilities with conversational prompts. Using \texttt{Llama2-7b-chat}~\citep{touvron2023llama2} on the WMT15 De->En test set with chunk size $n=5$, we tested both zero- and five-shot settings. As shown in Table \ref{tab:preliminary_study}, conversational \textsc{SimulMT} performed poorly even with 5-shot prompting. The failure analysis reveals that LLMs, trained primarily on complete sentences, struggle with partial source translation and tend to hallucinate completions when presented with fragments in a multi-turn dialogue format. To address this limitation, we propose to SFT LLMs on conversational \textsc{SimulMT} data. The following section details our approach to converting a normal bi-text corpus into conversational prompt format.

\input{figures/policy_generalization_example}
\subsection{SFT on Conversational \textsc{SimulMT} Data}\label{simul-data}
As conversational \textsc{SimulMT} data is not naturally available, we propose to synthesize \READ/ \WRITE chunks by segmenting sentence pairs from parallel corpora. Inspired by \citet{arthur-etal-2021-learning} which generates the oracle policy from word alignments, we further extend the approach by carefully addressing the impact of word reordering and improving the generalizability of the oracle policy. Specifically, we first build \emph{monotonic dependency graph} from the alignment of a sentence pair. We then segment the graph and convert these segments into \READ/ \WRITE pairs, followed by augmentation to improve its generalization across various latency demands (\Cref{fig:policy_generalization_example}). The process is explained below.

\input{sections/3.1-data-curation}

%% file: tables/tab-preliminary.tex
\begin{table}[t]
\centering
\resizebox{1.0\columnwidth}{!}{%
\begin{tabular}{@{}lllrr@{}}
\toprule
\multicolumn{2}{l}{Setting} & N-Shot & SacreBLEU & COMET \\ \midrule
\multicolumn{2}{l}{\textsc{OfflineMT}} & 0-Shot & 30.99 & 84.95 \\ \midrule
\multicolumn{2}{l}{Convers. \textsc{SimulMT}} & 0-Shot & 7.14 & 58.76 \\
\multicolumn{2}{l}{Convers. \textsc{SimulMT}} & 5-Shot & 13.51 & 69.03 \\ \midrule
\multicolumn{5}{c}{Convers. \textsc{SimulMT} 0-Shot Failure Case} \\ \midrule
Chunk 1 Input: & \multicolumn{4}{l}{Die Flugdaten zeigten, dass das} \\
Chunk 1 Response: & \multicolumn{4}{l}{\begin{tabular}[c]{@{}l@{}}The flight data showed that \textcolor{red}{the plane}\\ \textcolor{red}{was flying at an altitude of 35,000 feet.}\end{tabular}} \\ \midrule
Chunk 2 Input: & \multicolumn{4}{l}{Flugzeug auch bei einem zweiten} \\
Chunk 2 Response: & \multicolumn{4}{l}{\textcolor{red}{The plane was also flying during the second flight.}} \\ \midrule
Reference & \multicolumn{4}{l}{Flight data showed the plane had to pull out of second} \\ \bottomrule
\end{tabular}%
}
\caption{Performance comparison of \texttt{Llama2-7b-chat} on WMT15 De->En test set in zero-shot and few-shot conversational \textsc{SimulMT} settings. \textsc{OfflineMT} results are included as a baseline. The example failure case demonstrates how the LLM hallucinates completions (shown in \textcolor{red}{red}) when translating partial sentences, leading to compounding errors in subsequent chunks.
}
\label{tab:preliminary_study}
\end{table}

%% file: figures/policy_generalization_example.tex
\begin{figure*}
    \centering
    \resizebox{1.0\textwidth}{!}{
        \includegraphics{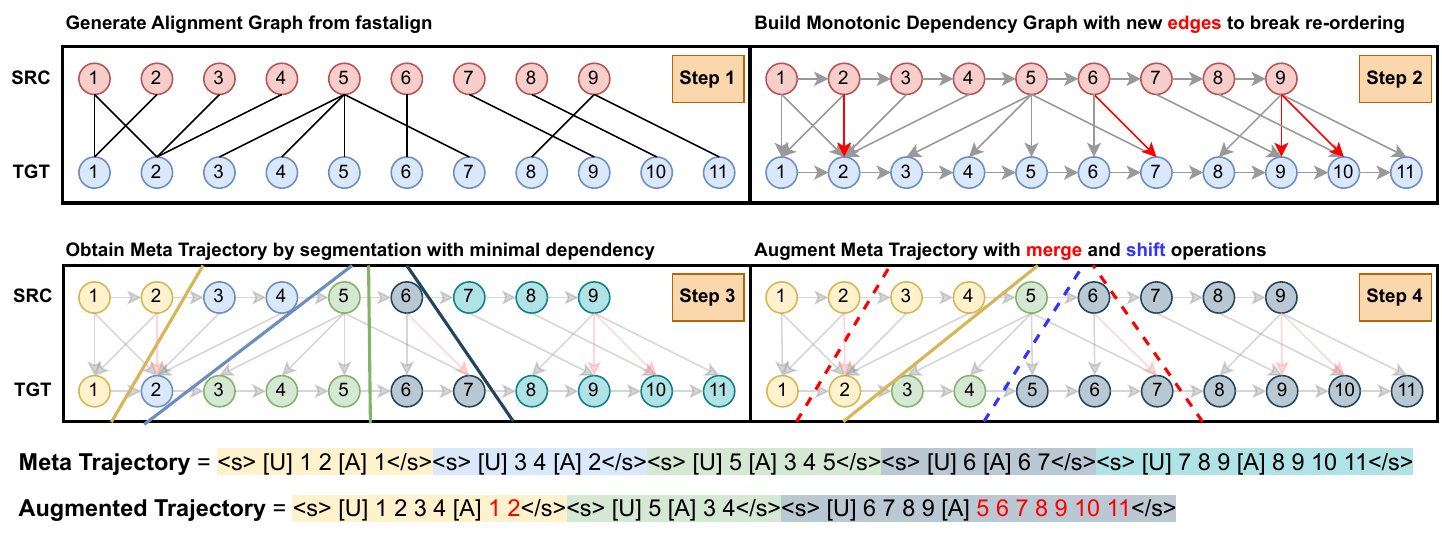}
    }
    \caption{The illustration of the data curating process. The first graph is obtained from \texttt{fast\_align}, it is then modified into a monotonic dependency graph by adding additional edges. The Meta Trajectory can be derived by segmenting the monotonic dependency graph with minimal dependency (segment with the colored solid line in step 3). Finally, Policy Generalization is applied to augment the segmented graph with merge (\textcolor{red}{red} dotted lines will be removed) and shift (\textcolor{blue}{blue} dotted lines are shifted) operations. Chunks in the trajectories derived from the third and fourth graphs are highlighted with different colors.}
    \label{fig:policy_generalization_example}
\end{figure*}

%% file: sections/3.1-data-curation.tex
\paragraph{Alignment Graph Generation} Given a sentence pair, we employ \texttt{fastalign}~\cite{fastalign} to obtain word alignment between source and target tokens (Step 1 in \Cref{fig:policy_generalization_example}). The obtained alignment is a set $\mathcal{A}$ of pairs $(i, j)$ denoting the source token $x_i$ is aligned with its corresponding target token $y_j$. We define the \emph{sufficient} source token set to generate a given target token $y_j$ as $\textbf{a}_{j} = \{i | (i, j) \in \mathcal{A}, \forall i \in [0, I]\}$.

A source and target sentences have a monotonic translation relationship if the previous target tokens only aligned with the previous source tokens, i.e. $\forall j > k \quad \min(\textbf{a}_j) \geq  \max(\textbf{a}_{k})$ \citep{koehn-etal-2005-edinburgh, ling-etal-2011-discriminative}. This condition ensures that the relative order of words is preserved between the source and target sentences. 
In that case, the optimal minimum-latency policy that retains sufficient source information is to produce the monotonic translation that follows the word order of the source sequence, i.e. \WRITE target token $y_j$ immediately after reading the final required source token $x_{\max(\textbf{a}_j)}$, and then \READ the next source tokens.

\paragraph{Monotonic Dependency Graph} Monotonic dependency enables effective implementation of optimal \READ/\WRITE policies. However, translations often require reordering to produce grammatically correct output, especially between languages with different syntactic structures. To address this, we propose constructing a monotonic dependency graph $\overrightarrow{\mathcal{A}}$ from alignment set $\mathcal{A}$ (Step 2 in \Cref{fig:policy_generalization_example}) such that the monotonic condition is met.

For each target token $y_j$ violating the monotonic condition $\min(\textbf{a}_j) < \max(\textbf{a}_{j-1})$, we add a new edge from the last sufficient source token $x_{\max(\textbf{a}_{j-1})}$ to $y_j$, eliminating the need for reordering. In \Cref{fig:policy_generalization_example}, $y_2$ violates monotonicity as its earliest required source token $\min(\textbf{a}_2) = 1$ precedes the last required source token for the previous target $\max(\textbf{a}_1) = 2$. Thus, we add an edge from $x_2$ to $y_2$.

\paragraph{Meta Trajectory} We then segment the monotonic dependency graph and convert these segments into \READ/ \WRITE pairs, representing the \emph{meta trajectory} of the oracle policy with minimum latency (Step 3 in Figure \ref{fig:policy_generalization_example}). We examine each target token to identify its exclusive corresponding source tokens with minimal dependency. Each subgraph $\overrightarrow{\mathcal{A}_{j}}$ corresponds to a pair $(R_j, W_j)$ where $W_j=\{y_j\}$ is a target token and $R_j=\{x_i | i \in \textbf{a}_{j} \setminus \textbf{a}_{j-1} \}$ contains new source tokens required since the previous target. 
When consecutive target tokens depend on the same source token, we combine their \WRITE actions, assigning the shared source token to $R_j=\{x_i\}$ and forming $W_j=\{y_j,...,y_{j+n}\}$. This generates a meta trajectory $RW^{\star} = [(R_1, W_1),...,(R_C, W_C)], C\leq I$, with $C$ chunks.

\paragraph{Trajectory Augmentation} Since the meta trajectories are tailored for minimal latency, they may not generalize well to different lengths of the input chunk, corresponding to different levels of latency. To improve the LLM's adaptability across various latency demands, we augment the meta-trajectory $RW^{\star}$ with a series of \textbf{merge} and \textbf{shift} operations (Step 4 in Figure \ref{fig:policy_generalization_example}).
We first traverse $RW^{\star}$ and randomly merge $\delta$ consecutive \READ and \WRITE actions, forming new pairs $([R_c,...,R_{c+\delta} ], [W_c,...,W_{c+\delta} ] )$, where $[\cdot]$ is the string concatenation operation. Here, $\delta$ is a variable re-sampled from a uniform distribution $\mathcal{U}(\delta_{\min}, \delta_{\max})$ where $\delta_{\min}$ and $\delta_{\max}$ are predefined hyperparameters.

Additionally, with a probability of $\beta$, we shift a portion of tokens from a \WRITE action $W_c$ to the next one $W_{c+1}$ in the merged trajectory. More specifically, we split $W_c$ at a proportion $\rho$ and transfer the latter part to the next pair, resulting in $(R_c, W_c^{<\rho}), (R_{c+1}, [W_c^{>\rho}, W_{c+1}])$ where $\rho$ is sampled from $\mathcal{U}(\rho_{\min},0.9)$ where $\rho_{\min}$ is a hyperparameter. 

This augmentation enhances the LLM's context conditioning and suits incremental decoding where prediction endings are often truncated by hypothesis selection algorithms. The resulting trajectory consists of \READ/\WRITE chunks of varying lengths, formatted with conversational prompts for SFT. During training, we apply cross-entropy loss only on target tokens within unshifted \WRITE chunks.

%% file: sections/4-experiment.tex
\section{Experiments}\label{sec:experiment}

\begin{table}[t]
\centering
\resizebox{\columnwidth}{!}{%
\begin{tabular}{@{}llccc@{}}
\toprule
\textbf{Trajectory} & \textbf{Dimension} & \textbf{De$\rightarrow$En} & \textbf{En$\rightarrow$Vi} & \textbf{En$\rightarrow$Zh} \\ \midrule
\multirow{3}{*}{\textbf{Meta-Trajectory}} & \textbf{\#Chunk} & $10.69\pm5.5$ & $12.98\pm8.1$ & $11.94\pm7.3$ \\
 & \textbf{\#SRC word/Chunk} & $1.74\pm0.8$ & $1.38\pm0.4$ & $1.68\pm0.5$ \\
 & \textbf{\#TGT word/Chunk} & $1.79\pm0.8$ & $1.73\pm0.5$ & $1.53\pm0.5$ \\ \midrule
\multirow{3}{*}{\textbf{Aug-Trajectory}} & \textbf{\#Chunk} & $2.74\pm1.2$ & $3.12\pm1.6$ & $2.95\pm1.4$ \\
 & \textbf{\#SRC word/Chunk} & $7.01\pm3.9$ & $5.83\pm2.8$ & $7.02\pm3.6$ \\
 & \textbf{\#TGT word/Chunk} & $7.18\pm3.9$ & $7.35\pm3.5$ & $6.40\pm3.2$ \\ \bottomrule
\end{tabular}%
}
\caption{Statistics of curated conversational \textsc{SimulMT} training data across all benchmarks, showing chunk counts and source/target tokens per chunk (mean$\pm$std) for both meta and augmented trajectories.}
\vspace{-1em}
\label{tab:trajectory_statistic}
\end{table}

\subsection{Datasets}
\label{app_sec:data_setting}
\paragraph{WMT15 De->En} (4.5M training pairs) We use \texttt{newstest2013} (3000 pairs) for validation and \texttt{newstest2015} (2169 pairs) for testing\footnote{\url{www.statmt.org/wmt15/}}. 

\paragraph{IWSLT15 En->Vi} (133K training pairs) We employ TED \texttt{tst2012} (1553 pairs) and \texttt{tst2013} (1268 pairs) as validation and test sets, respectively\footnote{\url{nlp.stanford.edu/projects/nmt/}}. 

\paragraph{MUST-C En->Zh}~\citep{di-gangi-etal-2019-must} (359k training pairs) This TED talk dataset provides 1349 pairs for validation and the \texttt{tst-COMMON} (2841 pairs) for testing.

\paragraph{Conversational \textsc{SimulMT} Datasets}
For each dataset, we create conversational prompt versions from their training sets using the approach described in \S\ref{simul-data}. We employ \texttt{fastalign}~\cite{fastalign} to obtain initial word alignment graphs. For trajectory augmentation, we set $\delta_{\min:\max}=(2,10)$ for merging operations. For shift operations, both $\beta$ and $\rho_{\min}$ are set to 0.5, meaning we shift at least 50\% of tokens in a target segment to the next one with 0.5 probability. \Cref{tab:trajectory_statistic} presents detailed statistics for these datasets.

\subsection{Evaluation Metrics} 
We evaluate translation quality and latency using SacreBLEU\footnote{BLEU+nrefs:1+case:mixed+eff:no+tok:\{13a,zh\}\\+smooth:exp+version:2.3.1}~\citep{matt-2018-sacrebleu}, COMET\footnote{\url{https://huggingface.co/Unbabel/wmt22-cometkiwi-da}}~\citep{rei2020comet}, and word-level average lagging (AL)~\citep{ma-etal-2019-stacl}. To assess computational efficiency, we measure word wall time (WWT)~\citep{wang2023simultaneous}, which represents the average time required to predict a word on identical hardware.

\subsection{Model Training} 
For all LLM-based methods, we use \texttt{Llama2-7b-chat}~\citep{touvron2023llama2} as the backbone following \citet{wang2023simultaneous}. We conduct QLoRA-based SFT~\cite{hu2022lora,dettmers2023qlora} for one epoch with $r=64$, $\alpha=16$, learning rate of 2e-4, batch size of 48, and 4-bit quantization on a single A100 GPU. Both offline and conversational prompt models are fine-tuned on identical data sources (standard offline style bitext from the aforementioned training sets), but formatted as offline prompts and conversational prompts respectively.

\input{figures/main_result}

\subsection{Settings}
\label{sec:settings}
We compare our proposed conversational \textsc{SimulMT} against the following baselines:

\paragraph{Encoder-Decoder Transformers}
We evaluate the performance of a series of specialized Encoder-Decoder Transformer models for both \textsc{OfflineMT} and \textsc{SimulMT}:
\begin{itemize}

\myitem \textbf{Offline NMT}: Following \citep{zhang2022itst}, we train vanilla Transformer~\cite{vaswani2017transformer} (48M parameters for En->Vi; 300M for De->En and Zh->En) with beam size 5 for inference.

\myitem \textbf{Wait-$k$} \citep{ma-etal-2019-stacl}: A fixed policy approach that reads $k$ source tokens before alternating read/write operations. We test with $k$ ranging from 1-8 for De->En and Zh->En, 4-8 for En->Vi.

\myitem \textbf{ITST}~\citep{zhang2022itst} An adaptive policy that measures the information transferred from source to target token and determines when to proceed with translation with a threshold (set as 0.1-0.7 for all datasets).

\myitem \textbf{Wait-Info} \citep{zhang-etal-2022-wait} An adaptive policy using token information thresholds ($\mathcal{K}$ from 1-8 for all datasets) to coordinate the timing of translation.

\end{itemize}

\paragraph{LLM-based \textsc{SimulMT}} We compare our conversational prompt approach against the offline prompt method~\citep{wang2023simultaneous}, using identical \READ policies with chunk sizes $n$=[3,5,7,9,11,13]. Both approaches are evaluated with RALCP hypothesis selection (beam=5). We also assess greedy decoding (beam=1, no hypothesis selection) with our conversational prompting only (as computational latency baseline), since offline prompting inherently requires hypothesis selection and cannot function with greedy search. For reference, we include results from LLM-based \textsc{OfflineMT} as a performance upper bound.

\subsection{Results}
Our preliminary study in \Cref{tab:preliminary_study} showed LLMs struggle with zero/few-shot conversational \textsc{SimulMT}. Here we examine whether fine-tuning on our curated data enables effective conversational \textsc{SimulMT}, focusing on quality-latency balance.
\input{figures/decoding_speed}

\paragraph{Translation Quality} As shown in \Cref{fig:main_result}, LLM-based approaches (\textcolor{red}{red} and \textcolor{blue}{blue}) outperform Transformer baselines (\textcolor{amber}{yellow}) across all language pairs by up to 3 BLEU/10 COMET points. With sufficient latency allowance, LLM-based \textsc{SimulMT} even surpasses offline Transformer NMT. At equivalent latency levels, our conversational prompting (\textcolor{red}{red}) achieves comparable BLEU scores to offline prompting (\textcolor{blue}{blue}) while often showing better COMET scores.

\paragraph{Translation Latency} Our conversational \textsc{SimulMT} (\textcolor{red}{red}) reduces latency compared to offline prompting (\textcolor{blue}{blue}), with average reductions of 1.17 and 1.50 AL across all benchmarks. For En->Vi and En->Zh, our approach achieves latency comparable to specialized \textsc{SimulMT} models. While RALCP (\textcolor{red}{\texttt{S:LLM-ConvPrompt-RALCP}}) generally provides better quality than greedy decoding (\textcolor{red}{\texttt{S:LLM-ConvPrompt-Greedy}}), the latter offers lower latency.

\paragraph{Practical Advantages} Most significantly, our conversational \textsc{SimulMT} (\textcolor{red}{red}) maintains superior translation quality at low latency levels (AL<4) compared to specialized models (\textcolor{amber}{yellow}), making it particularly valuable for practical applications requiring both high quality and low latency. In contrast, offline prompting (\textcolor{blue}{blue}) with identical decoding configurations struggles to operate effectively in the low-latency range, diminishing its quality advantages relative to specialized approaches (\textcolor{amber}{yellow}). These results demonstrate that our conversational prompting approach effectively addresses the efficiency-quality trade-off in simultaneous translation with LLMs.


%% file: figures/main_result.tex
\begin{figure*}
  \centering
    \includegraphics[scale=0.45]{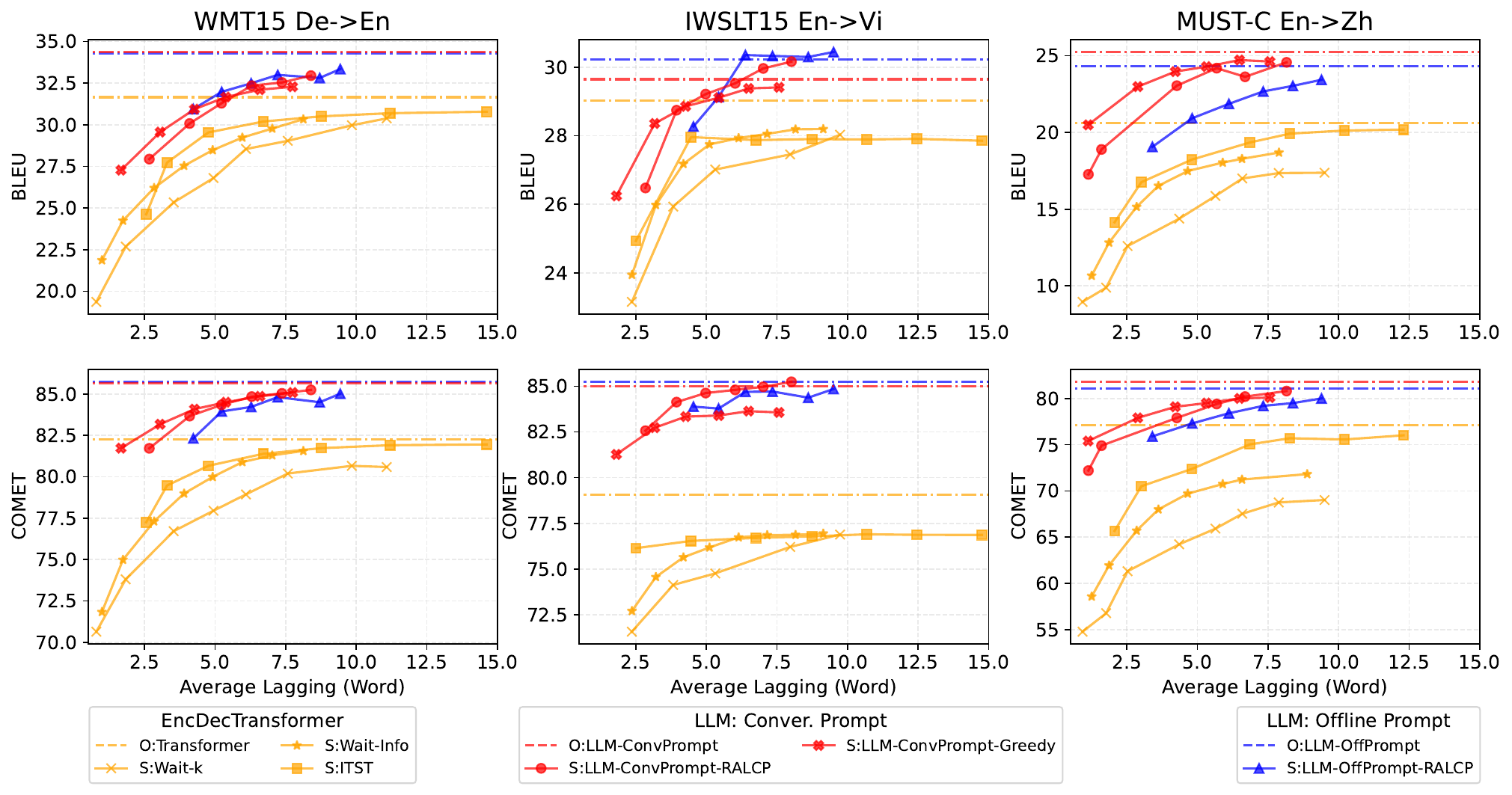}
  \caption{Translation quality and latency results on three benchmarks. Results are presented in three groups with different colors: (i) Encoder-Decoder Transformer baselines (\textcolor{amber}{orange}), (ii) Offline-Prompt LLMs (\textcolor{blue}{blue}), and (iii) Conversation-Prompt LLMs (\textcolor{red}{red}). Offline and Simultaneous decoding are distinguished by the first letter (O/S).}
  \label{fig:main_result} 
\end{figure*}

%% file: figures/decoding_speed.tex
\begin{figure*}
    \centering
    \resizebox{0.85\textwidth}{!}{
        \includegraphics{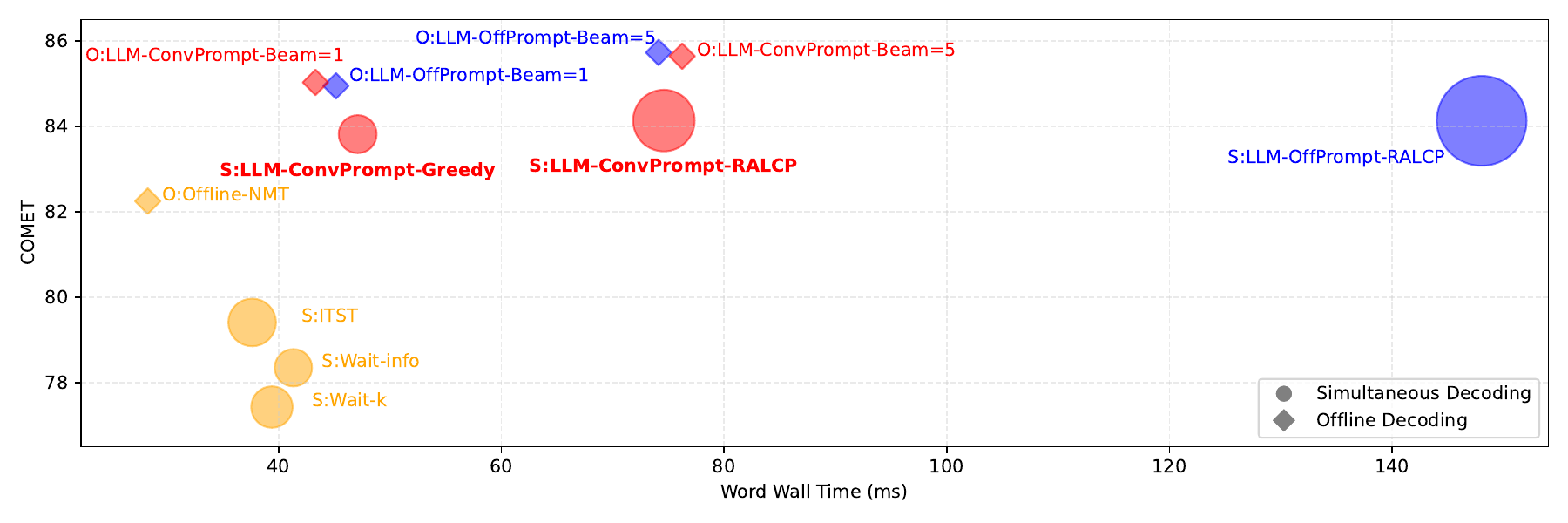}
    }
    \vspace{-3mm}
    \caption{Relationship between computational efficiency (Word Wall Time) and translation quality (COMET score) on WMT15 De->En. Simultaneous decoding settings are shown as circles, with circle size representing variance across different latency control parameters (e.g. $n$). Offline settings are represented by diamonds. Color coding matches \Cref{fig:main_result}, with our proposed approach highlighted in \textbf{bold}.}
    \label{fig:decoding_speed}
   \vspace{-1em}
\end{figure*}

%% file: sections/4.1-analysis.tex
\section{Analysis}

\subsection{Decoding Speed} 
While Average Lagging (AL) effectively quantifies algorithmic delay between translation and source input, it doesn't account for computational costs. In real-world applications, actual inference time critically impacts user experience: a model with low AL might still deliver poor user experience due to high computational overhead. To address this limitation, we evaluate decoding speed using Word Wall Time (WWT), which measures actual inference time per word (\S\ref{sec:settings}).

Figure~\ref{fig:decoding_speed} presents detailed WWT results for WMT15 De->En translation. Our analysis reveals that offline prompting with RALCP (\textcolor{blue}{\texttt{S:LLM-OffPrompt-RALCP}}) exhibits the slowest performance, making it impractical despite good translation quality. In contrast, our conversational prompting approach with RALCP (\textcolor{red}{\texttt{S:LLM-ConvPrompt-RALCP}}) achieves computational efficiency comparable to offline LLM translation (\textcolor{red}{\texttt{O:LLM-ConvPrompt-Beam=5}}) while maintaining high translation quality. 

Most notably, our conversational prompting with greedy decoding (\textcolor{red}{\texttt{S:LLM-ConvPrompt-Greedy}}) delivers the best efficiency-quality balance—achieving processing speeds comparable to specialized \textsc{SimulMT} models (\textcolor{amber}{yellow}) while producing significantly better translations. These results demonstrate that our approach effectively addresses both algorithmic and computational latency concerns, making it suitable for practical deployment.

\input{figures/policy_generalization}
\subsection{Effectiveness of Trajectory Augmentation}
\vspace{-0.5em}
To evaluate our trajectory augmentation strategy, we conducted an ablation study comparing models trained on: (\emph{i}) meta trajectories only, (\emph{ii}) meta trajectories with merge operations, and (\emph{iii}) meta trajectories with both merge and shift operations (\S\ref{simul-data}). All models used identical hyperparameters, with training data as the only variable.

As shown in \Cref{fig:policy_generalization}, trajectory augmentation yields notable improvements in translation quality and latency when using RALCP. The merge operation contributes most significantly to these improvements, while models trained solely on meta trajectories perform poorly across all metrics.

This suggests augmentation techniques enhance the model's ability to generalize across different latency conditions. Without augmentation, the model struggles with varying input chunk sizes, causing RALCP to accept less reliable hypotheses and increasing latency. The augmented approach effectively prepares the model for dynamic simultaneous translation scenarios.

\label{app_sec:context_following}
\input{figures/context_following}

\subsection{Ability to Leverage Contextual Information}
Effective \textsc{SimulMT} with conversational prompting requires the model's ability to accurately utilize contextual information. To evaluate this capability, we designed an experiment isolating the model's performance on the final chunk of translation both with and without access to preceding context.

For each test instance, we extracted the complete inference history and separated it into: (i) the source-target dialogue history serving as context, and (ii) the final source chunk representing the latest input. We then tasked our fine-tuned LLM with translating this final chunk under two conditions: with and without access to the preceding conversation history. Performance was evaluated by computing BLEU scores on the concatenation of the generated final chunk with its original history.

As shown in \Cref{fig:context_following}, we observed a consistent 2-point decrease in BLEU scores when context was withheld. This performance gap demonstrates our model effectively leverages information from previous conversation turns to produce more accurate translations, confirming the fine-tuned LLM maintains translation coherence.

\subsection{Generalizability Across LLM Families}
\input{figures/llm_variation}

In our main experiments, we used \texttt{Llama-2-7b-chat} following \citet{wang2023simultaneous} for consistency. Now, we examine our approach's generalizability across different LLMs, using identical training and inference parameters for fair comparison. We report only greedy simultaneous decoding and offline beam=5 results to eliminate interference with hypothesis selection.

\paragraph{Impact of Model Iteration} We compare \texttt{Llama-2-7b-chat} with the newer \texttt{Llama-3.1-8B-Instruct}~\citep{grattafiori2024llama3herdmodels} on WMT15 De->En to assess how model advancements affect performance. As shown in \Cref{fig:llm-generation}, the newer model demonstrates consistent improvements in both offline and simultaneous modes. This confirms that conversational \textsc{SimulMT} effectively transfers to newer LLMs, with benefits from improved instruction-following capabilities and enhanced language modeling.

\paragraph{Effect of Model Scale} We investigate how model size impacts performance by comparing \texttt{Llama-3.1-8B-Instruct} with the smaller \texttt{Llama-3.2-3B-Instruct}~\citep{grattafiori2024llama3herdmodels} on WMT15 De->En. \Cref{fig:llm-scale} shows that while the larger model predictably outperforms its smaller counterpart, the 3B model still achieves acceptable translation quality (on par with \texttt{Llama-2-7b-chat} in \Cref{fig:llm-generation}), suggesting our method is viable on resource-constrained devices.

\paragraph{Impact of Target Language Proficiency} We evaluate \texttt{Llama-3.1-8B-Instruct} against \texttt{Qwen2.5-7B-Instruct}~\citep{qwen2025qwen25technicalreport} on MUST-C En->Zh to investigate the effect of the model's target language capabilities. As shown in \Cref{fig:llm-language}, Qwen2.5 consistently outperforms Llama-3.1 for Chinese translation by 1-2 BLEU points across all latency settings, demonstrating that target language proficiency provides additional benefits with our approach.

%% file: figures/policy_generalization.tex
\begin{figure}
    \centering
    \resizebox{0.9\columnwidth}{!}{
        \includegraphics{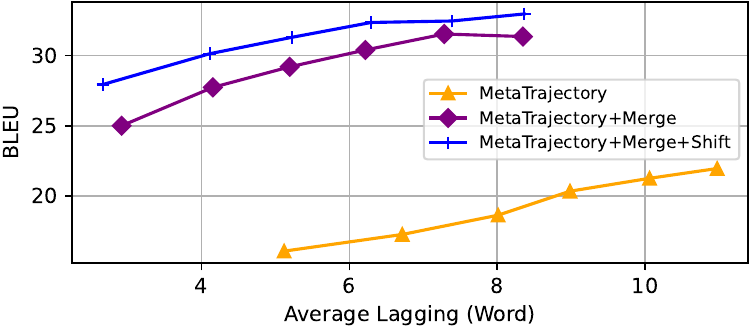}
    }
     
    \vspace{-3mm}
    \caption{Effect of trajectory augmentation strategies on translation quality (BLEU) and latency (AL) for WMT15 De->En. Results compare models trained on meta-trajectories alone versus with merge and shift operations.}
    \label{fig:policy_generalization}
   \vspace{-2em}
\end{figure}

%% file: figures/context_following.tex
\begin{figure}
    \centering
    \resizebox{1\columnwidth}{!}{
        \includegraphics{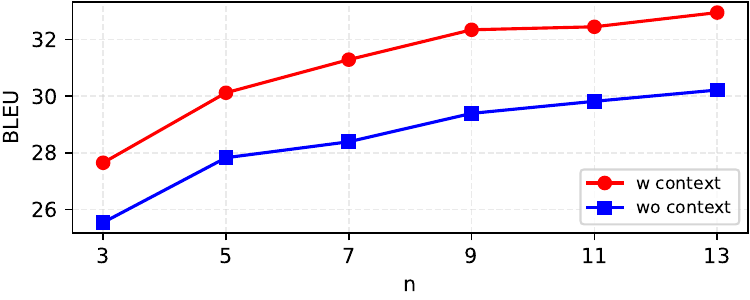}
    }
    \vspace{-5mm}
    \caption{Translation quality (BLEU) on WMT15 De->En when generating the final chunk with vs. without preceding context, across different chunk sizes. The consistent gap demonstrates effective context utilization.}
    \vspace{-1em}
    \label{fig:context_following}
\end{figure}

%% file: figures/llm_variation.tex
\begin{figure}[t]
    \centering
    \begin{subfigure}[t]{0.5\textwidth}
    \centering
    \footnotesize
    \includegraphics[width=1.0\textwidth]{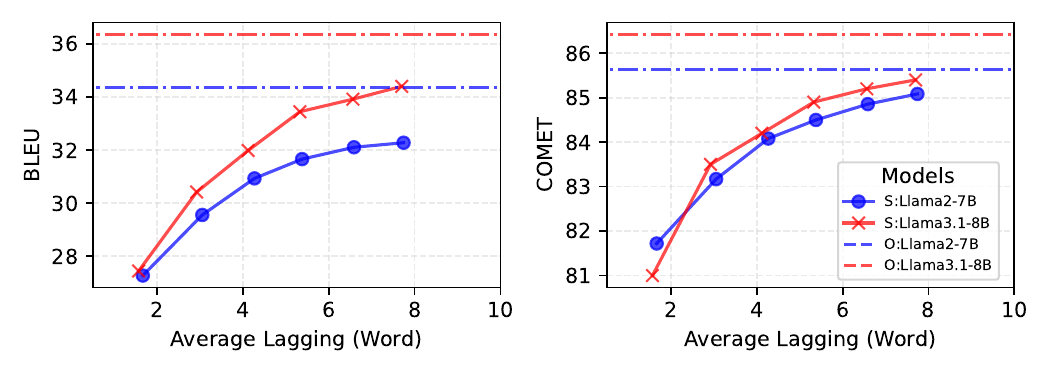}
    \vspace{-1.0em}
    \caption{Impact of model iteration (Llama-2-7b-chat vs. Llama-3.1-8B-Instruct) on WMT15 De->En.}
    \label{fig:llm-generation}
    \end{subfigure}%
    \hfill
    \vspace{-0.2em}
    \begin{subfigure}[t]{0.5\textwidth}
    \centering
    \footnotesize
    \includegraphics[width=1.0\textwidth]{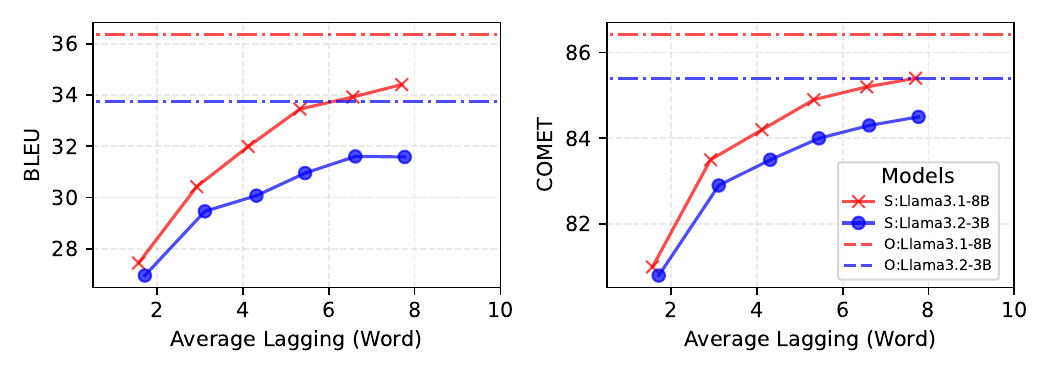}
    \vspace{-1.0em}
    \caption{Effect of model scale (Llama-3.1-8B-Instruct vs. Llama-3.2-3B-Instruct) on WMT15 De->En.}
    \label{fig:llm-scale}
    \end{subfigure}%
    \hfill
    \vspace{-0.2em}
    \begin{subfigure}[t]{0.5\textwidth}
    \centering
    \footnotesize
    \includegraphics[width=1.0\textwidth]{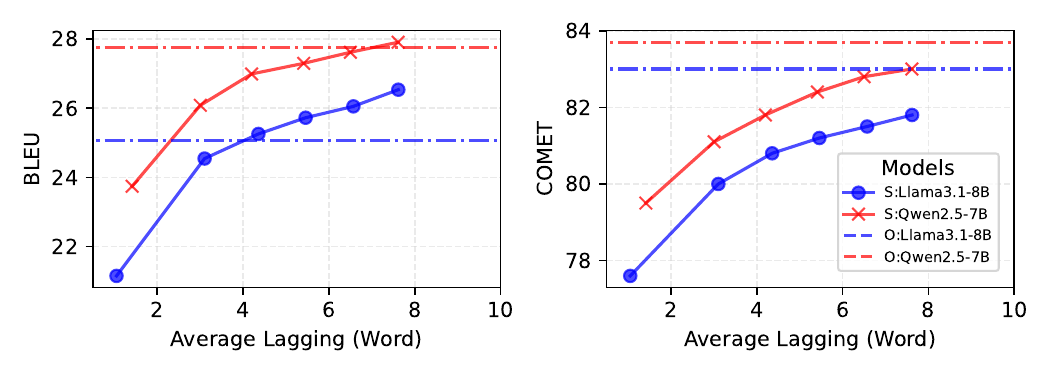}
    \vspace{-1.0em}
    \caption{Impact of target language proficiency (Llama-3.1-8B-Instruct vs. Qwen2.5-7B-Instruct) on MUST-C En->Zh.}
    \label{fig:llm-language}
    \end{subfigure}%
    \vspace{-0.2em}
    
    \caption{Performance comparison of different LLM families with our conversational prompt.}
    \vspace{-0.4em}
    \label{fig:llm-variation}
\end{figure}

%% file: sections/5-related-work.tex
\section{Related Works}
\paragraph{Simultaneous Machine Translation (\textsc{SimulMT})} is the task to provide real-time translation of a source sentence stream where the goal is to minimize the latency while maximizing the translation quality. A common approach is to train an MT model on prefix-to-prefix dataset to directly predict target tokens based on partial source tokens~\citep{ma-etal-2019-stacl}. Alternatively, \citet{low_latency} proposed the incremental decoding framework to leverage the pretrained \textsc{OfflineNMT} model and turn it into a \textsc{SimulMT} model without further training. A core component of \textsc{SimulMT} is a read-write policy to decide at every step whether to wait for another source token (\texttt{READ}) or to generate a target token (\texttt{WRITE}). Previous methods have explored fixed policy, which always waits for $k$ tokens before generation ~\citep{ma-etal-2019-stacl, zhang-etal-2022-wait} and adaptive policy, which trains an agent via reinforcement learning~\citep{gu-etal-2017-learning, arthur-etal-2021-learning}.  Re-translation~\citep{DBLP:conf/acl/ArivazhaganCMCY19} from the beginning of the source sentence at the \texttt{WRITE} step will incur high translation latency. Stable hypothesis detection methods such as Local Agreement, hold-$n$~\citep{low_latency} and Share prefix SP-n~\citep{nguyen21c_interspeech} are employed to commit stable hypothesis and only regenerate a subsequence of source sentence. The goal is to reduce the latency and minimize the potential for errors resulting from incomplete source sentence~\citep{polak-etal-2022-cuni,wang2021diformer}. 

\paragraph{LLM-based NMT} Recent research has delved into the potential usage of LLMs in MT~\cite{DBLP:journals/corr/abs-2302-09210,DBLP:journals/corr/abs-2304-04675,DBLP:journals/corr/abs-2309-07423}, especially in handling discourse phenomena~\citep{wang-etal-2023-document-level, wu2024adapting} and linguistic nuances such as idioms~\citep{manakhimova-etal-2023-linguistically} and proverbs~\citep{wang2025proverbsrunpairsevaluating}.
While LLMs do exhibit some level of translation capability, prior research has identified that they still lags behind the conventional NMT models, especially for low resource languages~\citep{DBLP:journals/corr/abs-2309-07423}. Additionally, the translation performance varies depending on prompting strategies~\citep{Zhang2023PromptingLL}.
Efforts have been made to enhance the LLMs' MT performance by incorporating guidance from dictionary~\citep{lu2023chainofdictionary}, further fine-tuning~\citep{zeng2023tim,xu2023paradigmshift} and augmenting with translation memories~\citep{mu-etal-2023-augmenting}. 


\paragraph{LLM-based \textsc{SimulMT}} SimulLLM \citep{agostinelli2023simul} explore the ability to adapt an LLM finetuned on NMT task to simultaneous translation with wait-k strategy. \citet{wang2023simultaneous} adopt hybrid READ/WRITE policy with wait-k and incremental decoding. TransLLaMA \citep{koshkin2024transllama} teach LLMs to produce WAIT tokens to preserve the causal alignment between source and target tokens. At each inference round, LLMs only produce a single word or WAIT token, which is very costly due to multiple rounds of LLM calls. 
\citet{guo2024sillm} introduce LLM into the \textsc{SimulMT} task as a translation agent working with a specialized \textsc{SimulMT} policy agent. An additional memory module stores translation history. The policy agent decides on \texttt{READ}/\texttt{WRITE} actions, while the LLM translates target segments. They face the same KV-cache reuse challenge noted by \citet{wang2023simultaneous}, making the computational cost of collaborating big and small models even more significant.

%% file: sections/6-conclusion.tex
\section{Conclusion}


This paper focuses on the feasibility of utilizing LLM for \textsc{SimulMT}. We found that leveraging the incremental-decoding framework with offline prompting leads to high computational latency, hindering the reuse of the Key-Value cache. To address this, we propose the conversational prompting which allows LLMs to conduct \textsc{SimulMT} in a multi-turn dialogue manner. The approach significantly speeds up the inference and also preserves the quality superiority, enabling practical LLM-based \textsc{SimulMT} systems.


%% file: sections/7-limitations.tex
\section*{Limitations}
We summarize the limitations of this study in the following aspects:
\vspace{-1.5mm}
\paragraph{Data} Our evaluation was conducted on three commonly used benchmarks which may limit the diversity in domains, styles, and languages. There may also be potential data contamination concerns since LLMs might have been exposed to parts of our test sets during pre-training. A more comprehensive evaluation with diverse datasets across more domains and language pairs would strengthen our findings.

\vspace{-1.5mm}
\paragraph{Alignment-based Data Curation} Our approach relies on word alignment tools like \texttt{fast\_align} to segment parallel sentences, which has inherent limitations. These tools may struggle with languages having drastically different word orders or grammatical structures, potentially creating suboptimal segmentation points. Furthermore, the alignment quality degrades for distant language pairs or complex sentences with idiomatic expressions and cultural references. While our augmentation strategies help mitigate some issues, they are still constrained by the initial alignment quality. 

\section*{Ethics Statement}
Our work is built on top of an existing LLM. For this reason, we share the similar potential risks and concerns posed by the underlying LLM. Our method is trained on commonly used training resources of the Machine Translation research community and as such we are not expecting our approach to introduce new areas of risks.

%% file: sections/8-appendix.tex
\clearpage

\section*{Appendix}
\label{sec:appendix}
\appendix

\section{Conversational SimulMT Decoding}
\label{sec:appx-algo}

\input{sections/resources/algo_policy}
\Cref{algo:simul_decode} presents the details of applying conversational prompts for decoding.

%% file: sections/resources/algo_policy.tex
\begin{algorithm}[t]
\begin{algorithmic}[1]
    \REQUIRE LLM : $\texttt{LLM}_{\theta}$, \\
    Source chunks: $\vx=[]$, \\
    Target chunks: $\vy=[]$, \\
    KV-Cache: $\mathbf{h}=[]$, \\
    Chunk index: $c=0$, \\
    Variables Definition: Source chunk size: $n$, Beam-size: $B$, Agreement-degree: $\gamma$ \\
    \WHILE{NOT\_FINISH}
        \STATE $\vx_c \gets \texttt{READ($n$)}$ //\texttt{READ} $n$ tokens
        \STATE $\vx$.append($\vx_c$)
        \STATE $\vx_{\text{prompt}} \gets \texttt{PROMPT}(\vx, \vy)$
        \STATE $\vy_c^{\prime},\mathbf{h}^{\prime} \gets \texttt{LLM}(\mathbf{x}_{\text{prompt}}, B, \mathbf{h}, \text{latest=True})$
        \STATE //$B$ candidates with latest tokens in $\vy_c^{\prime}$
        \STATE $\vy_c, \mathbf{h} \gets \texttt{PREFIX}(\vy_c^{\prime}, \mathbf{h}^{\prime})$
        \STATE //Prune with Prefix selection, e.g. RALCP
        \IF{$\vy_c == \emptyset$}
            \STATE \textbf{continue}
        \ELSE
            \STATE $\vy$.append($\vy_c$)
            \STATE $\texttt{WRITE}(\vy_c)$
            \STATE $c \gets c+1$
        \ENDIF
    \ENDWHILE
\end{algorithmic}
\caption{Conversational \textsc{SimulMT} Decoding}
\label{algo:simul_decode}
\end{algorithm}